\documentclass[letterpaper,10pt]{article} 

\usepackage{opticameet3}

\usepackage{subcaption}
\newcommand\authormark[1]{\textsuperscript{#1}}
\newcommand{\figref}[2]{Fig.#1#2}

\usepackage{svg}
\usepackage{amsmath,amssymb}
\usepackage[colorlinks=true,bookmarks=false,citecolor=blue,urlcolor=blue]{hyperref} 
\usepackage{textpos} 

\begin{document}

\begin{textblock*}{\textwidth}(0cm, -2cm) 
\footnotesize
© 2024 IEEE. Personal use of this material is permitted. Permission from IEEE must be obtained for all other uses, in any current or future media, including reprinting/republishing this material for advertising or promotional purposes, creating new collective works, for resale or redistribution to servers or lists, or reuse of any copyrighted component of this work in other works.
\rule{\textwidth}{0.4pt}
\end{textblock*}

\vspace{-9pt}
\title{Hardware-In-The-Loop Training of a 4f Optical Correlator with Logarithmic Complexity Reduction for CNNs}


\author{Lorenzo Pes \authormark{1*}, Maryam Dehbashizadeh Chehreghan \authormark{1}, Rick Luiken \authormark{1},  Sander Stuijk \authormark{1}, Ripalta Stabile \authormark{1}, Federico Corradi \authormark{1}}

\address{\authormark{1} Technische Universiteit Eindhoven, Den Dolech 2, 5612 AZ, Eindhoven}

\email{\authormark{*}l.pes@tue.nl} 

\begin{abstract}
This work evaluates a forward-only learning algorithm on the MNIST dataset with hardware-in-the-loop training of a 4f optical correlator, achieving 87.6\% accuracy with $O(n^2)$ complexity, compared to backpropagation, which achieves 88.8\% accuracy with $O(n^2 \log{n})$ complexity.
\end{abstract}

\section{Introduction}

Optical computing offers additional degrees of parallelism compared to digital computing by leveraging multiplexing across wavelength, polarization, and spatial modes. It also benefits from the superior propagation speed and lower energy dissipation of photons compared to electrons, making it a promising approach for accelerating Artificial Intelligence (AI) computations beyond the capabilities of conventional electronic systems \cite{shastri_photonics}. Convolutional Neural Networks (CNNs) are a fundamental tool in various AI image processing tasks. Nevertheless, CNNs face computational challenges in high-resolution image processing (e.g., medical imaging) with a computational complexity of $O(n^2k^2)$ for an $(n \times n)$ image and a $(k \times k)$ kernel. Performing convolutions in the Fourier domain transforms them into point-wise multiplications. Optical devices like 4f correlators naturally execute Fourier transforms using lenses and perform multiplications with Spatial-Light-Modulators (SLMs), offering a $O(1)$ complexity for the convolution operation, making them highly efficient for large-scale computations \cite{miscuglio2020}.

Despite these advantages, the analog nature of optical hardware introduces imperfections that are difficult to model in software-based training, leading to performance gaps in deployment. Hardware-in-the-loop (HWL) training integrates the physical hardware directly into the training loop, allowing models to adapt to device-specific characteristics \cite{spall2022hybrid}. However, HWL approaches using error backpropagation (BP), which requires a digital model for gradient computation, limits the full exploitation of the photonic substrate. In contrast, forward-only (FO) learning algorithms \cite{kohan_sp,hinton_ff,giorgia_pepita,pau_mempepita} rely solely on the forward pass outputs for parameter updates. 

The characterization of HWL training using FO algorithms in a 4f optical correlator has not yet being addressed. Thus, with this work, we explore HWL training of a 4f optical correlator using both error backpropagation and one prominent FO algorithm, the PEPITA algorithm \cite{giorgia_pepita}, evaluating their training and inference performance. We demonstrate that PEPITA eliminates the need for software-based gradients, allowing models to adapt to device-specific characteristics while reducing the complexity from $O(n^2 \log{n})$ to $O(n^2)$. 

\section{Forward-only training of optical devices}
Recent implementations of forward-only (FO) algorithms in optical systems \cite{xue_ff_opt, oguz_ff_opt, momeni2023backpropagation} have primarily focused on the Forward-Forward (FF) algorithm \cite{hinton_ff}, demonstrating its effectiveness in optical training. However, other FO algorithms, such as Signal-Propagation (SP) \cite{kohan_sp}, PEPITA \cite{giorgia_pepita}, and its memory-efficient variant MEMPEPITA \cite{pau_mempepita}, have also emerged, offering diverse approaches beyond FF. Even though FF is model-agnostic and avoids activation storage, it requires multiple forward passes during inference for classification tasks, making it impractical for efficient deployments. In contrast, SP also avoids activation storage and uses only one forward pass during inference but requires model knowledge, making it less suitable for physical devices. PEPITA requires only one forward pass during inference and is model-agnostic, but it requires activation storage like BP. MEMPEPITA, however, reduces memory usage by performing an extra forward pass during training to re-compute activations, eliminating the need for activation storage. This work characterizes PEPITA due to its equivalent performance to MEMPEPITA while requiring only two forward passes, optimizing computation over memory.

The computational graph for HWL training of a convolution layer using both BP and PEPITA is depicted in \figref{1}{c} and \figref{1}{d} respectively. We define $W$ as the convolution learnable parameter, $n^2$ the size of input image $I(x,y)$. Additionally, we define $z_{hw}$ as the optical convolution output and $h$ as the output of the activation function. Notably, PEPITA does not require a differentiable model of the device, but instead, it uses activations from two forward passes ($h$ and $h_{mod}$) to compute the parameter update $\Delta W$, directly leveraging the physical device and making the update local in space, with a total computational complexity of $O(n^2)$. In contrast, to calculate the parameter update $\Delta W$ with BP, the derivative $\partial z_{hw} / \partial W$ requires a differentiable software model of the convolution, with a total complexity of $O(n^2 \log{n})$ due to the use of FFT and point-wise multiplications. Additionally, the final layer error $e^L$ must be back-propagated through the layers to compute $\partial L / \partial z_{hw}$. Thus, PEPITA provides a logarithmic reduction in computation over BP while removing the need for device knowledge, making it an efficient solution for photonic device training.

\begin{figure}[t!]
    \centering
    \includegraphics[width=1\textwidth]{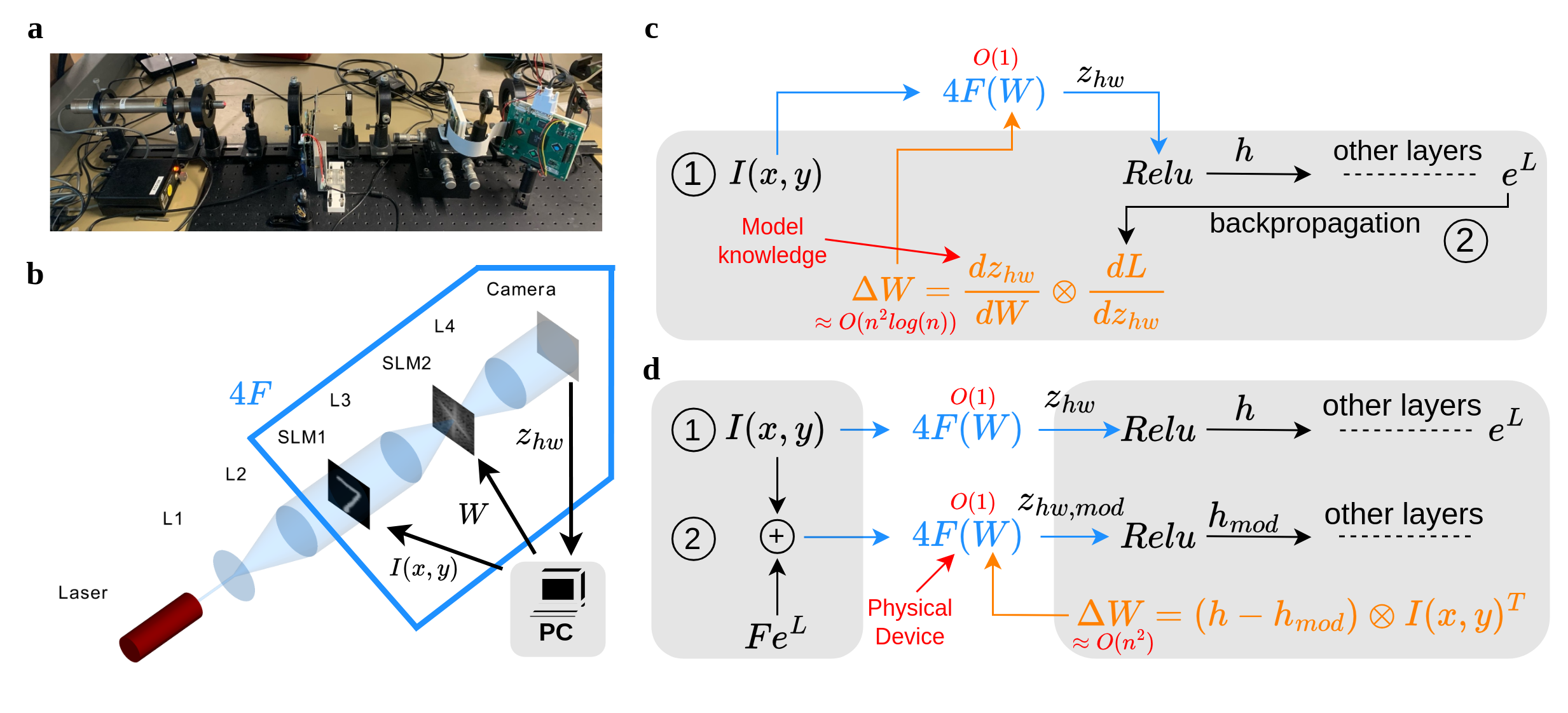}
    
    \caption{\footnotesize \textbf{HWL with 4f Optical Correlator}. \textbf{a} Physical 4f correlator. \textbf{b} Full system view with input and output processing. \textbf{c} HWL training with BP. \textbf{d} HWL training with PEPITA. In both \textbf{c} and \textbf{d}, the blue path represents the physical device. In red are the computational complexity of the update rules represented by the orange equations. }
    \label{fig:correlator}
\end{figure}

\section{Setup and experiment}

In our experiments, we use the 4f optical correlator of \figref{1}{a}. A pictorial representation of the full system is provided in \figref{1}{b}. It consists of two 8-bit SLMs. A 633nm HeNe laser beam is expanded through lenses L1 and L2 to fully illuminate SLM1, where the input image $ I(x, y) $ is encoded. The modulated light passes through lens L3, producing the Fourier transform $ \mathcal{F}\{ I(x, y) \} $ at the focal plane, where SLM2 applies a point-wise multiplication with a kernel $W$. To exploit the convolution theorem, lens L4 then performs the inverse Fourier transform, $ \mathcal{F}^{-1}\{ \mathcal{F}\{ I(x, y) \} \cdot W \} $, returning the signal to the spatial domain as the convolution $ I(x, y) * W$, which is captured by a high-speed camera. Parameter updates are computed in the spatial domain, while the kernel is represented in the frequency domain inside the device. Hence, the updated kernel $W' = W - \eta \Delta W$, where $\eta$ is the learning rate, is transformed to the Fourier domain using the Fast Fourier Transform (FFT).

To evaluate the performance of traditional error BP and the PEPITA forward-only algorithm in adapting to hardware imperfections, we perform HWL using a sub-sample of the MNIST dataset, consisting of gray-scale images of handwritten digits. Given the optical device's slow sequential processing time, limited by the 25ms setup time of the SLMs, we train on 600 samples and test on 100. The model architecture includes a single Fourier convolution layer with 8 kernels of size $28 \times 28$, followed by a max-pooling layer ($2 \times 2$), and a fully connected classification layer. During HWL training and testing, the convolution operation is executed on the optical device. Each experiment is repeated with 5 different seeds for statistical validity. No extensive hyperparameter tuning was conducted; all experiments used a learning rate $\eta = 0.001$ with the Adam optimizer and a camera exposure of 20ms.

\section{Results and Discussion}
The experimental results are summarized in \figref{2}{}. Both algorithms minimize classification loss effectively, with BP and PEPITA converging to similar values. During testing, BP shows a slight generalization advantage ($88.8 \pm 4 $) over PEPITA ($87.6 \pm 3 $), likely due to limited hyperparameter tuning. Notably, PEPITA has a $\log{n}$ reduction in computational complexity. The Structural-Similarity-Index-Metric (SSIM) between software and optical convolutions has a mean of $\approx 0.8$, highlighting a notable discrepancy; however, the training still achieves competitive performance. To fully leverage the optical device's potential and enable analysis of the entire dataset, parallelism must be introduced. Encoding multiple input images on SLM1 and multiple kernels on SLM2 would significantly enhance throughput beyond the current sequential processing. Currently, the SLM setup time of 25 ms limits processing to 40 images per second, excluding overhead from the operating system and camera exposure time. Directly interfacing the SLMs and camera with a Field-Programmable-Gate-Array (FPGA) would reduce overhead, maximizing processing efficiency and realizing the device’s full capability.

\section{Conclusion}
In conclusion, this work demonstrates that forward-only hardware-in-the-loop training using the PEPITA algorithm can effectively train optical correlators for CNN applications, achieving comparable performance to BP, with a reduced computational overhead of $\log{n}$. Further optimization, such as parallelism and FPGA integration, is necessary to fully exploit the potential of the optical device and enable large-scale dataset analysis.

\section{Acknowledgment}
This publication is funded in part by the project NL-ECO: Netherlands Initiative for Energy-Efficient Computing (with project number NWA. 1389.20.140) of the NWA research programme Research Along Routes by Consortia which is financed by the Dutch Research Council (NWO).  

\begin{figure}[t!]
    \centering
    \includegraphics[width=\textwidth]{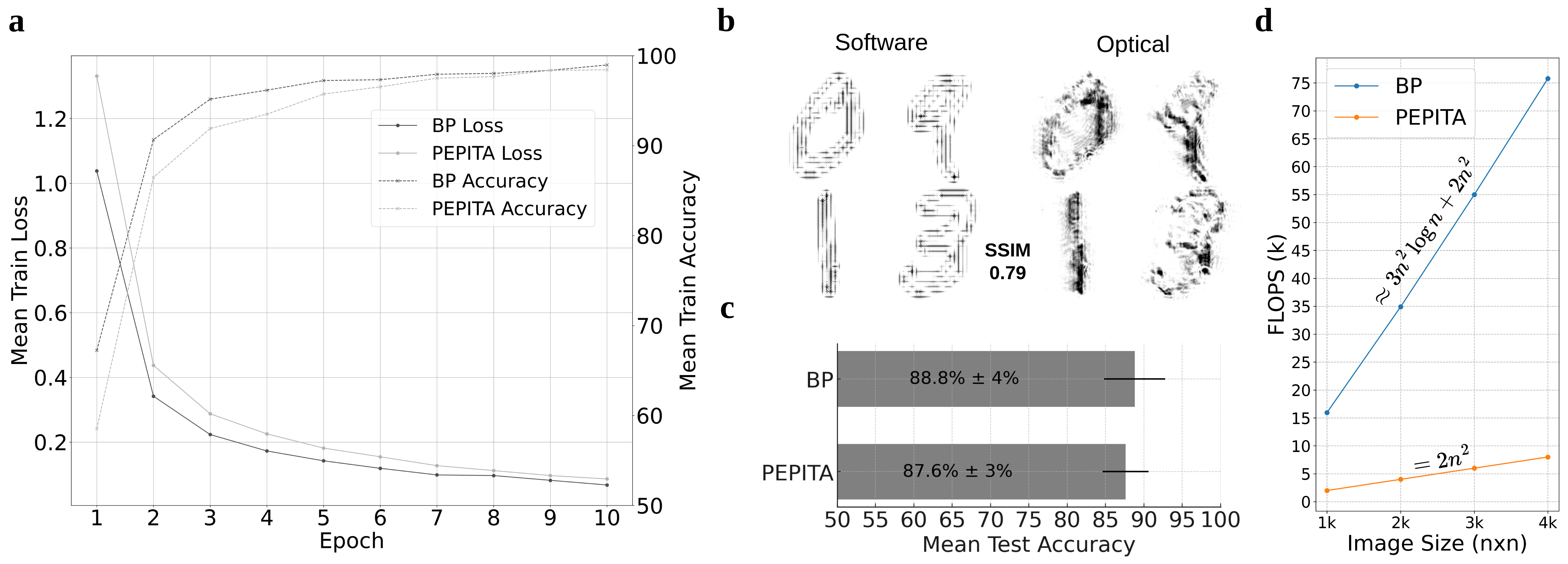}

    \caption{ \footnotesize \textbf{HWL vs. software training results}. \textbf{a} Mean training loss and train accuracy per epoch for each algorithm with HWL with 300 MNIST samples. \textbf{b}  Software versus (experimental) optical convolution with an edge detection kernel. The average SSIM is reported inside. \textbf{c} End of training mean test accuracy in optical device. \textbf{d} Total FLOPS for BP and PEPITA. Approximate FLOPS for BP are computed without accounting for $dL/d z_{hw}$ from downstream layers, which would further increase the total FLOPS. }
    \label{fig:results}
\end{figure}

\begingroup
\footnotesize  
\bibliographystyle{unsrt}

\endgroup

\end{document}